\documentclass[10pt,twocolumn,letterpaper]{article}

\usepackage{wacv}
\usepackage{times}
\usepackage{epsfig}
\usepackage{graphicx}
\usepackage{amsmath}
\usepackage{amssymb}

\usepackage{algorithm}
\usepackage{amsmath}
\DeclareMathOperator*{\argmin}{argmin}

\usepackage[numbers]{natbib}

\usepackage{bm}

\usepackage[pagebackref=true,breaklinks=true,letterpaper=true,colorlinks,bookmarks=false]{hyperref}

\wacvfinalcopy 

\def\wacvPaperID{582} 

\ifwacvfinal\fi
\begin{document}

\title{A Generative Approach to Zero-Shot and Few-Shot Action Recognition}

 \author{ Ashish Mishra$^{*,1}$ , Vinay Kumar Verma$^{\dag,1}$, M Shiva Krishna Reddy$^{*}$, Arulkumar S$^*$\\   Piyush Rai$^\dag$ and Anurag Mittal$^*$\\
{ $^*$Indian Institute of Technology Madras \qquad $^\dag$Indian Institute of Technology Kanpur}\\
{\tt\small \{vkverma,piyush\}@cse.iitk.ac.in, \{mishra,shiva,aruls,amittal\}@cse.iitm.ac.in}
}
\maketitle
\ifwacvfinal\thispagestyle{empty}\fi

\begin{abstract}
   We present a generative framework for zero-shot action recognition where some of the possible action classes do not occur in the training data. Our approach is based on modeling each action class using a \emph{probability distribution} whose parameters are functions of the attribute vector representing that action class. In particular, we assume that the distribution parameters for any action class in the visual space can be expressed as a linear combination of a set of basis vectors where the combination weights are given by the attributes of the action class. These basis vectors can be learned solely using labeled data from the known (i.e., previously seen) action classes, and can then be used to predict the parameters of the probability distributions of unseen action classes. We consider two settings: (1) \textit{Inductive setting}, where we use only the labeled examples of the seen action classes to predict the unseen action class parameters; and (2) \textit{Transductive setting} which further leverages \textit{unlabeled data} from the unseen action classes. Our framework also naturally extends to few-shot action recognition where a few labelled examples from unseen classes are available. Our experiments on benchmark datasets (UCF101, HMDB51 and Olympic) show significant performance improvements as compared to various baselines, in both standard zero-shot (disjoint seen and unseen classes) and generalized zero-shot learning settings. 
   
\end{abstract}

\section{Introduction}
\footnotetext[1]{Both authors contributed equally.}
Action Recognition is an important problem in Computer Vision in which knowledge about a sequence of actions is learned from a large collection of video clips.  It is a challenging task due to the inherent variability in actions, non-deterministic occlusion patterns, abrupt changes in illumination, cluttered dynamic background, and noisy videos. Knowledge about an action is inferred usually by learning from the labelled data in a supervised manner.  Even as more complex models are being built, it is a common observation that the number of categories of actions is progressively increasing (for example, one of the earliest benchmark datasets KTH has 6 categories while Olympic, HMDB and UCF datasets have 16, 51, and 101 categories, respectively).  Consequently, annotating videos of this growing number of categories can be a very cumbersome task and consequently restricts the scalability of a fully supervised action recognition for a large number of categories.

\begin{figure*}[!htbp]
\begin{center}
\includegraphics[height=7cm, width=17cm]{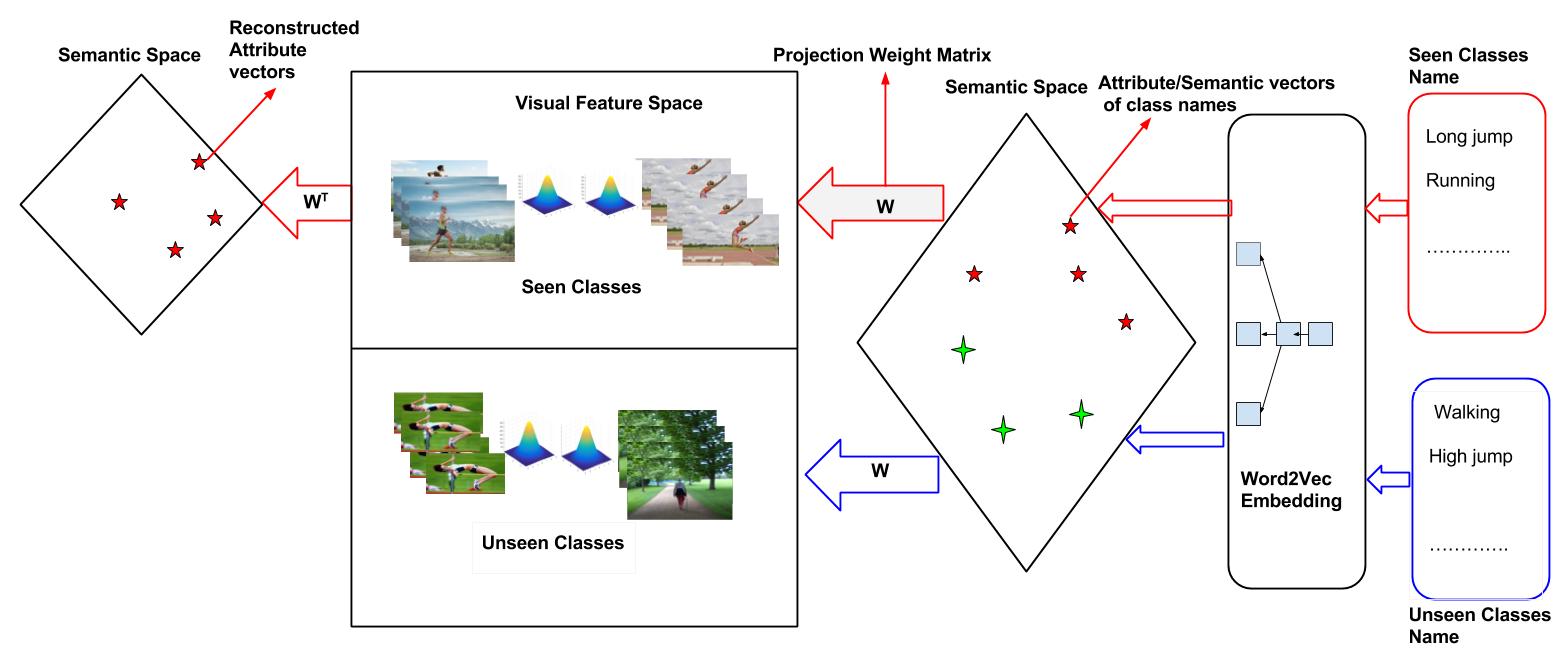}
\end{center}
   \caption{\textbf{Proposed Model}: Each class attribute is projected to the visual space, In the visual space each class is represented by a Gaussian distribution. To avoid information loss, a reconstruction regularizer is added.}
\label{fig:short}
\end{figure*}

To circumvent this problem, Zero-Shot Learning (ZSL) of actions has been actively pursued~\cite{xu2016multitask-zero,xu2017transductive,qin2017zero}.  In the conventional Action Recognition framework, only the classes present in the training data can be recognized by the model during the test phase. In Zero-Shot Learning, however, the model is expected to recognize and categorize action classes that did not appear in the training phase at all. The information about the unseen classes is provided via other modalities such as language in the form of textual descriptions, \textit{word2vec} \cite{word2vec} or human annotated attributes.  Essentially, the model has to learn to recognize the unseen action classes based on the knowledge acquired from the data instances of the seen action classes. Zero-shot learning is typically defined in two settings: (1) the conventional setting, in which set of classes for the training and test instances are disjoint ($Y_{tr}\cap T_{te}=\varnothing$); and (2) the generalized zero-shot (GZSL) setting, in which the set of classes for the training and test instances may have an overlap~\cite{HAA,ConSE}. The generalized zero-shot setting is considered much harder than standard setting (disjoint setting) since the learned models tend to be biased towards predicting seen classes at training time (as they are learned solely from the unseen class training data). While much of the prior work in ZSL has focused on the conventional setting, the focus has recently shifted to the more realistic GZSL setting. 


In this work, we present a simple generative approach for zero shot action recognition, which works in both standard as well as generalized ZSL setting. Our approach models each action class as a probability distribution in the visual space where the parameters of this distribution are assumed to be a linear combination of a set of ``basis'' parameters, where the combination weights are given by the (known) attribute vector of that class. This is akin to assuming that each action class can be represented as a combination of a set of ``prototype'' action classes. The complete architecture of our model is shown in Fig.~\ref{fig:short}.

Once the basis vectors are learned using the training instances from seen action classes, the parameters of an unseen action class distribution can be easily computed via a weighted combination of the learned basis vectors, with weights being the attributes of the respective unseen class. The loss function (More details in Methodology section) is formulated in such a way that, for each seen action class, the weighted combination of the basis vectors is close to the maximum likelihood estimate (MLE) of the class distribution's parameters. The MLE estimate can also be replaced by maximum-a-posteriori (MAP) estimate. Our approach is akin to the one proposed recently in~\cite{verma2017simple} for zero-shot learning, though our focus an application is specifically the zero-shot \emph{action recognition} problem. In addition, we add an additional ``reverse direction'' regularizer to encourage reconstruction ability of the class attribute vectors from the parameters of the seen class distributions so as to minimize the information loss. Note that this is akin to an autoencoder (cf, Fig.~\ref{fig:short}). One of the appealing aspects of our model is that it admits a simple closed-form solution.

Our main contributions can be summarized as follows
\begin{itemize}
 \item We provide a probabilistic generative approach for zero-shot learning (ZSL) where each action class is represented by a Gaussian distribution (although the Gaussian can be replaced by other distributions without changing the rest of our model).
 \item We show that our approach, although simple, generalizes well to the unseen classes in the inductive setting and improves over the state-of-the-art.
 \item We show that our approach can be easily generalized to the transductive setting where unlabeled data from unseen classes are available at training time.
 \item Our model also naturally extends to be ``few-shot learning'' setting where a few examples of each unseen class are available as well. In particular, the parameters of the class distribution can be updated easily given a few additional labeled examples from that class. Through extensive experimentation on three benchmark datasets, we show that our simple approach gives significant performance gains in all three settings over the state-of-the-art methods. 
 \item Finally, since our approach is generative, we can also synthesize novel examples for any unseen class by sampling from the respective class distribution. Since we can now have labeled data from seen as well as unseen classes, it is possible to train a classifier in the generalized zero-shot setting which is much harder than the standard (disjoint) setting.
\end{itemize}

\section{Methodology}\label{sec:methodology}

For the zero-shot action recognition setting, we denote the total number of seen action classes by $S$ and the total number of unseen action classes by $U$. We take a generative classification based approach to the action recognition problem where we assume that the data instances of each action class (seen/unseen) $c$ are generated by a distribution $p(\mathbf{ x}|\bm{\theta_c})$. Without loss of generality, and for simplicity of exposition, we will assume these distributions to be Gaussians (note that our approach can be used with other distributions as well). In the Gaussian case, the parameters \bm{$\theta_c$} consist of the mean vector $\bm{\mu_c} \in \mathbb{R}^D$ and a diagonal covariance  matrix $\bm{\Sigma}_c = \text{diag}(\bm{\sigma_c^2})$, where $\bm{\sigma_c^2} \in \mathbb{R}_+^D$. We assume a diagonal covariance matrix to reduce the total number of parameter estimated and prevent overfitting especially when the number of examples from each class is small. However, other forms for the covariance matrix can also be used. 

Given labeled data from the seen classes, it is straightforward to estimate the parameters $\bm{\mu_c},\bm{\sigma_c}$ using Maximum Likelihood Estimation (MLE) or Maximum-a-Posteriori (MAP) estimation. For example, using MLE, the mean is estimated as $\bm{\mu_{c}}=\frac{1}{N_c}{\sum_{i=1}^{N_{c}}\mathbf{x_i}}$ and $\bm{\sigma_{c}^2}=\text{diag}(\frac{1}{N_c}{\sum_{i=1}^{N_{c}}(\mathbf{x_i}-\bm{\mu_{c}})(\mathbf{x_i}-\bm{\mu_{c}})^\top})$ where $N_c$ denotes the number of labeled examples from class $c$.

However, this approach cannot be used to estimate the parameters $\bm{\theta_c}~(c=S+1,...,S+U)$ of unseen classes due to unavailability of labeled data corresponding to unseen classes. To resolve this problem, we model the parameters $\bm{\theta_c} = (\bm{\mu_c},\bm{\sigma_c^2})$ of each seen/unseen clas  as a function of the respective class attribute vector $\mathbf{ a_c}$, i.e., $\bm{\theta_c} = f(\mathbf{a_c})$. In the zero-shot learning setting, the class attribute vector $\mathbf {a_c} \in \mathbb{R}^K$ is either provided by a human expert or as the \textsc{word2vec} embedding of the name of the action.

The function $f$ can be linear or nonlinear and can be learned using the labeled data instances of seen classes in visual feature space. Once learned, the function $f$ can be used to predict $\bm{\theta_c}$ for all the unseen class actions $c=S+1,\ldots,S+U$ using their respective class attributes. 

A simple choice of $f$ is a linear model that maps the class attributes $\mathbf{a_c}$ to the class parameters $\bm{\theta_c}$. In the Gaussian class distribution case, for the mean $\bm{\mu}_c$, such a linear function $f$ can be defined as

\begin{equation}
    \bm{\mu_{c}} = f_{\mu}(\mathbf{a_{c}}) = \mathbf{W_{\mu}a_{c}}
\end{equation}
Note that the above linear model represents the mean $\bm{\mu}_{c} \in \mathbb{R}^{D}$ as a weighted linear combination of $K$ basis vectors $\mathbf{W_\mu} =\mathbf{[w_{\mu_1},w_{\mu_2},..,w_{\mu_{K}}]} \in \mathbb{R}^{D\times K}$ is a set of \textit{learned} basis vectors in the visual space. 

The basis vectors $\mathbf{W_\mu}$ can be learned using the seen class training data. In particular, given the empirical estimates $\bm{\hat{\mu}_c}, c=1,\ldots,S$ of means of the seen class distributions, we can use $(\mathbf{a_c},\bm{\hat{\mu}_c})$ as ``training data'' to learn the regression model $\mathbf{W_\mu}$ that maps $\mathbf{a_c}$ to $\bm{\hat{\mu}_c}$.

While the above model can be seen as mapping the class attribute vector $\mathbf{a_{c}}$ to the class mean $\bm{\mu}_{c}$, we further impose the condition that the class means can also be used to \emph{reconstruct} the class attribute vector via a ``reverse map'' akin to an autoencoder, i.e., $\mathbf{a_{c}} = \mathbf{W_{\mu}^T \bm{\mu}_{c}}$, which leads to
\begin{equation}
 \bm \mu_c = \mathbf{W_\mu a_c} = \mathbf{W_\mu W_\mu^\top \bm \mu_c}
\end{equation}

A similar procedure can be employed for learning the mapping from the class attributes $\mathbf{a_{c}}$ to the variance parameters $\bm{\sigma}_c^2$ of the distribution of class $c$ via another set of basis vectors $\mathbf{W_{\sigma^2}}$. Sections~\ref{sec:linreg} and~\ref{sec:nonlinreg} provide more details.
 Once the basis vectors $\mathbf{W_\mu},\mathbf{W_{\sigma^2}}$ (which define the functions $f_\mu$ and $f_\sigma$) are learned, we can use them to estimate the parameters (e.g., $\bm \mu_c,\bm \sigma_c^2$) of the distribution of each unseen class. For example, given the class attribute vector $\mathbf{a_{c}}$ of an unseen class $c=S+1,\ldots,S+U$, we can estimate $\bm{\mu_{c}}$ simply as $\bm{\mu_{c}} = \mathbf{W_{\mu}a_{c}}$.
 
 The mapping $f$ (which is essentially a regression model) from the class attribute vector to the parameters of the class distribution can be linear or nonlinear. We describe both these cases in the next two sections.
 
\subsection{Linear Regression} 
\label{sec:linreg}
Given the labeled data from seen classes $c=1,...,S$, we can estimate their class distribution parameters using MLE. We can then learn the functions $f_{\mu}$ and $f_{\sigma^{2}}$ using training data of the form $\mathbf{(a_c,\bm \mu_c)}_{c=1}^S$ and $\mathbf{ (a_c,\bm \sigma^{2}_{c})}_{c=1}^S$. In the linear regression approach $\bm \mu_c = f_\mu(\mathbf{a_c})$ and $\bm \sigma_c^2 = f_{\sigma}(\mathbf{a_c})$, we assume the functions $f_\mu$ and $f_\sigma$ to be linear projections with weight matrices, $\mathbf{W}_\mu$ and $\mathbf{W}_\sigma$, making this problem equivalent to the following regression problem:
\begin{equation*}
\bm \mu_{c}=\mathbf{W_{\mu}a_c}  \quad \text{ s.t. } \quad \mathbf{a_c=W_{\mu}^T\bm \mu_{c}}
\end{equation*}
\begin{equation*}
\begin{aligned}
\rho_c=\log \bm \sigma^{2}_{c}=\mathbf{W_{\sigma^{2}}a_c}  \quad \text{ s.t. } \quad \mathbf{a_c=W_{\bm \sigma^{2}}^T\bm \sigma^{2}_{c}}\\
\end{aligned}
\end{equation*}

The projection matrices $\mathbf{W_\mu}$ and $\mathbf{W_{\sigma^2}}$ can be easily learned using a multi-output ridge regression problem with training data $\mathbf{(a_c,\bm \mu_c)}_{c=1}^S$ and $\mathbf{(a_c,\bm \sigma^{2}_{c})}_{c=1}^S$. These problems have simple closed form solution and we omit the equations here for brevity. We give details equations for the nonlinear case, as shown below.

\subsection{Nonlinear Regression}
\label{sec:nonlinreg}
For the non-linear regression, we first map the attributes $\{\mathbf{a_c\}}_{c=1}^S$ to the kernel space using the kernel function $k$ which is defined as a nonlinear mapping $\phi$. Using the Representer theorem \cite{kernellearning}, we can re-formulate the regression problem in kernel space as given in Eq.~\ref{kernel_regression}. Note that instead of computing the $\phi(\mathbf{ a_c })$ explicitly, we have to compute only the dot product $\phi(\mathbf{a_c})^T\phi(\mathbf{a_{c'}})=k(\mathbf{a_{c}},\mathbf{a_{c'}})$ for the non-linear mapping of the two class $c$ and $c'$. 
Let $\mathbf{ K}$ be the kernel matrix of size $S\times S$ containing pairwise similarities of the attributes of the seen classes, $\mathbf{M}$ be the $D\times S$ matrix containing the means of the distributions of all the seen classes, then the attribute to mean nonlinear mapping can be learned by solving the following problem
\begin{equation}
\label{kernel_regression}
\begin{aligned}
\min_{\mathbf{W}_{\mu}}||\mathbf{M - W_{\mu}K||} ^{2}_{F} +\lambda_{\mu}||\mathbf{W}_{\mu}||^{2}_{2} \\ \text{s.t.} \quad
\mathbf{ K=W_\mu^*M}
\end{aligned}
\end{equation}

Eq \ref{kernel_regression} shows our main objective function. Here the first term can be interpreted as learning an optimal weight matrix that projects the attribute space to the visual space using the kernel regression. The second term ensures that we can reconstruct the attribute vector from the visual space and acts as a regularization term. Akin to an autoencoder~\cite{SAE2017}, we assume the two mappings to be reverse of each other

\begin{equation*}
\mathbf{ W_\mu^*=W}_\mu^T
\end{equation*}
Therefore the complete objective can be written as:
\begin{equation}\label{objectivefun}
\begin{aligned}
\mathbf{ W}^{*}_{\mu}=\argmin_{\mathbf{W}_{\mu}}||\mathbf{M - W_{\mu}K}||^{2}_{F}+\lambda_{\mu}||\mathbf{W}{\mu}||^{2}_{2}\\
+\lambda_{1}||\mathbf{K-W_{\mu}^{T}M}||_{F}^{2}
\end{aligned} 
\end{equation}
The next section provides details of the optimization procedure used for solving Eq.~\ref{objectivefun}
\subsubsection{Optimization}
Noting $Tr(\mathbf{K})=Tr(\mathbf{K}^T)$ and $Tr(\mathbf{W}_{\mu}^{T}\mathbf{ M})=Tr(\mathbf{ M}^T\mathbf{ W}_\mu)$, Eq.~\ref{objectivefun} can be written as:

\begin{equation}\label{eq13}
\begin{aligned}
\mathbf{W}^{*}_{\mu}= \argmin_{\mathbf{W}_{\mu}}||\mathbf{M - W_{\mu}K}||^{2}_{F}+\lambda_{\mu}||\mathbf{W}{\mu}||^{2}_{2}\\
+\lambda_{1}||\mathbf{K}^T-\mathbf{M}^T\mathbf{W}_\mu||_{F}^{2}
\end{aligned} 
\end{equation}
Taking the derivative of Eq.~\ref{eq13} and equating to zero we have.
\begin{equation}
\mathbf{ MM}^{T}\mathbf{W}_{\mu}+\mathbf{ W}_{\mu} \lambda_{1}\mathbf{ KK}^{T}+\lambda_{\mu}\mathbf{ W}_{\mu}=(1+\lambda_{1})\mathbf{ MK}^{T}
\end{equation}
\begin{equation}
\mathbf{MM}^{T}\mathbf{W_{\mu}+W}_{\mu} (\lambda_{1}\mathbf{KK}^{T}+\lambda_{\mu}) =(1+\lambda_{1})\mathbf{MK}^{T}
\end{equation}
The above equation has the form

\begin{equation}
\mathbf{ AW+WB=C}
\end{equation}
This is a well-known Sylvester equation which can be solved using the Bartels-Stewart algorithm \cite{sylbester} efficiently, and several off-the-shelf solvers exist (we used a MATLAB implementation for the same). The various quantities in the above equation are defined as
\begin{equation}
\mathbf{ A=MM}^{T}
\end{equation}
\begin{equation}
\mathbf{ B}=\lambda_{1}\mathbf{KK}^{T}+\lambda_{\mu}
\end{equation}
\begin{equation}
\mathbf{ C}=(1+\lambda_{1})\mathbf{MK}^{T}
\end{equation}
Likewise, the nonlinear model $f_{\sigma^{2}_i}$ can be learned by solving:
\vspace{-5pt}
\begin{equation}\label{sigma_opt}
\begin{aligned}
\mathbf{ W}^{*}_{\sigma^{2}}=\argmin_{\mathbf{ W }_{\sigma^{2}}}||\mathbf{R - W_{\sigma^{2}}K}||^{2}_{F}+\lambda_{\sigma^{2}}||\mathbf{ W}_{\sigma^{2}}||^{2}_{2} \\ +\lambda_{2}||\mathbf{ K-W}_{\sigma^{2}}^{T}\mathbf{ R}||_{F}^{2}
\end{aligned} 
\end{equation}

Again, taking derivatives and setting to zero gives
\begin{equation}
\mathbf{RR}^{T}\mathbf{W}_{\sigma^{2}}+\mathbf{W}_{\sigma^{2}} (\lambda_{2}\mathbf{KK}^{T}+\lambda_{\sigma^{2}}) =(1+\lambda_{2})\mathbf{RK}^{T}
\end{equation}
The above equation is also in the form of $\mathbf{ AW+WB=C}$
\begin{equation}
\mathbf{A=RR}^{T}
\end{equation}
\begin{equation}
\mathbf{B}=\lambda_{2}\mathbf{ KK}^{T}+\lambda_{\sigma^{2}}
\end{equation}
\begin{equation}
\mathbf{C}=(1+\lambda_{2})\mathbf{ RK}^{T}
\end{equation}
Given the learned parameters $\mathbf{W}_{\mu_c}$ and $\mathbf{W}_{\sigma^{2}_c}$, the parameters of data distribution for unseen classes $c=S+1,\dots,S+U$ are estimated as:
\begin{equation}
\bm \mu_c =\mathbf{ W_{\mu}k_c}, \quad \& \quad \bm \sigma^{2}_c =\exp(\rho_c)=\exp(\mathbf{ W_{\sigma^{2}}k_c)}
\end{equation}
Where $\mathbf{ k_c=[k(a_c,a_1),...,k(a_c,a_S)]}$ denotes an $S\times 1$ vector of kernel-based similarities of the class attribute vectors of the unseen class $c$ with the class attribute vectors of all the seen classes.

In the aforementioned procedure for estimation of the unseen class distribution parameters uses only seen class labelled data. In this setting, the unseen classes unlabeled data have not been used. This setting is called as an \emph{inductive setting}. If we have access to the unseen classes test instances at the training time, we can use these to improve the estimation of distribution parameters of unseen classes. This is the \emph{transductive setting} which we describe next.

\subsection{Transductive setting}\label{sec:transductive}
\begin{figure}[t]
\begin{center}
\includegraphics[height=4cm, width=8.6cm]{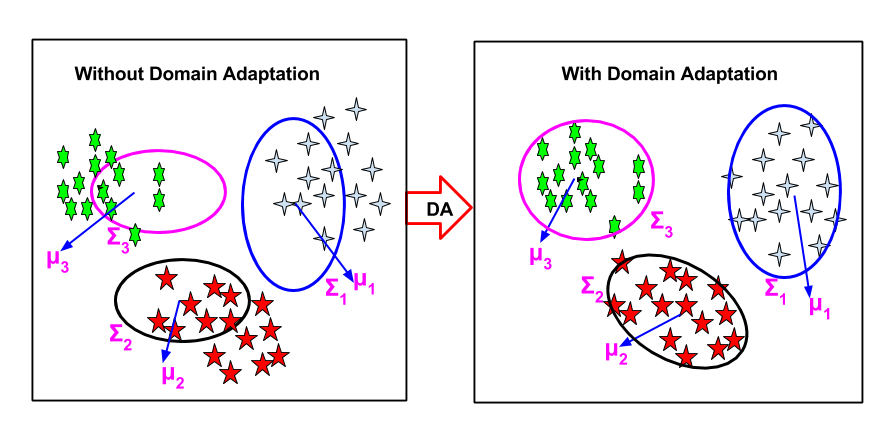}
\end{center}
\caption{\textbf{Domain Adaptation illustrative example}: Each class attribute is projected to the visual space, In the visual space each class are represented by a distribution. Because the seen and unseen class are disjoint, there is a problem of domain shift.}
\label{fig:tra}
\end{figure}

One of the unique advantages of the proposed generative approach is that unlabeled data from unseen classes can be leveraged to improve the parameter estimates ($\bm \mu_c$ and $\bm \sigma_c$). In zero-shot learning, training and test data could possibly come from different domains. Therefore, it is very likely that parameters learned in the training, will not work well for the test data. This phenomenon is called domain shift. An illustrative view of the domain shift can be seen in Fig.~\ref{fig:tra}. One way to overcome this issue is to use unlabeled data to further fine-tune the parameters learned by the inductive approach which only uses the labeled data from the seen action classes. In the transductive setting~\cite{xu2017transductive}, we assume that the test data is also available at the training time. This data can help mitigate the bias towards the seen classes. In this work, we handle the domain shift problem by initializing the parameters $\bm \mu_c,\bm \sigma_c$ using the learned basis vectors from the inductive learning phase, which are then fine-tuned using the unlabeled test data from the unseen classes using the an Expectation-Maximization (EM) algorithm.

Since each class distribution is assumed to be a Gaussian, this EM based procedure is equivalent to a Gaussian mixture model (GMM) on the unlabeled test data $(\mathbf{ x_n})_{n=1}^{N_u}$ from unseen classes. This GMM has $U$ mixture components, with each corresponding to an unseen class and is initialized by the estimated parameters of unseen classes $(\bm \mu_c,\bm \sigma^{2}_{c})_{c=S+1}^{S+U}$  in the inductive setting. 
The procedure for transductive setting can be briefly summarized as follows
\begin{enumerate}
\item \textbf{Initialize:} Let the initial estimate of the unseen class parameters be $\bm \Theta=(\bm \mu_c,\bm \sigma^{2}_{c})_{c=S+1}^{S+U}$ where $\bm \mu_c = \mathbf{ W_\mu a_c}$, $\bm \sigma^2_c = \exp \left(\mathbf{W_{\sigma^2} a_c}\right)$. Here $\mathbf{W}_\mu$ and $\mathbf{ W}_{\sigma^2}$ are estimated from seen class data using equations \ref{objectivefun}, \ref{sigma_opt} (assuming we have used the nonlinear regression model in the inductive phase). 
\item \textbf{Expectation Step:} Infer the probabilities for each example $\mathbf{x_n}$ belonging to each of the unseen classes $c=S+1,...,S+U$ as
\begin{equation*}
p(y_n=c|\mathbf{ x_n},\bm \theta)\propto \mathcal{N}(\mathbf{x_n}|\bm \mu_c,\bm \sigma^{2}_c)
\end{equation*}
where the class priors $p(c)$ are assumed to be uniform.
\item \textbf{Maximization Step:} Use the inferred class labels to re-estimate $\Theta=(\bm \mu_c,\bm \sigma^{2}_{c})_{c=S+1}^{S+U}$. These updates have closed form solution as in the standard GMM.
\item Go to step 2 if not converged.
\end{enumerate}

\subsection{Few-shot Action Recognition}
In few-shot action recognition, we have a small number of labeled examples for each of the unseen classes. Since our method assumes a Gaussian distribution for each class, we can easily extend our zero-shot action recognition method to few-shot action recognition. To this end, we treat the initial estimate obtained using the previous approach as the prior. Due to the conjugate nature of the Gaussian, we can update the estimates $(\mu_c,\sigma^{2}_{c})_{c=S+1}^{S+U}$ obtained from zero-shot action recognition method in a straightforward manner when such labeled data for unseen classes is provided. In particular, given a small number of labeled data $(\mathbf{ x_n})_{n=1}^{N_c}$ for unseen class $c$ the parameters of this class can be directly updated as:
\begin{equation}
\bm \mu^{FS}_c = \frac{\bm \mu+\sum_{n=1}^{N_c}\mathbf{ x_n}}{1+N_c}
\label{fs1}
\end{equation}
\begin{equation}
\bm \sigma^{2(FS)}_c=(\frac{1}{\bm \sigma^{2}_c}+\frac{N_c}{\bm \sigma^{2*}})^{-1}
\label{fs2}
\end{equation}
where $\bm \sigma^{2*}=\frac{1}{N_c}\sum_{n=1}^{N_c}(\mathbf{x_n}-\bm \mu_c)^{2}$ denotes empirical variance of $N_c$ observations from the unseen class $c$.

\subsection{Extension to Other Distributions}

Finally, we would like to emphasize that although we have consider Gaussians to model each class, our approach applies to any parameteric distribution $p(\mathbf{x}|\theta_c)$ as it essentially boils down to learning a mapping from the class attribute vectors $\mathbf{a}_c$ to the distribution parameters $\theta_c$. The choice of a Gaussian or an exponential family distribution~\cite{verma2017simple}, due to conjugacy, makes our estimation procedure particularly simple in the transductive and few-shot settings, but our framework is not restricted to these. Other density estimation methods such as deep generative models can also be used~\cite{aaai18}.

\section{Related Work}
ZSL can be viewed as an interplay of three subproblems: a visual representation of data instances (feature representation), semantic representation of all classes such as \textit{word2vec} representation \cite{word2vec}, and learning a function which establishes the relationship between visual representations and semantic representations of each class\cite{DAP,IAP}.

For visual (or feature) representation of class instance, popular hand-crafted features such as HOG \cite{HOG}, HOF \cite{HOF}, ITF \cite{ITF} were designed.  However, the proven utility of deep features for many tasks such as Object Recognition \cite{alexnet,vggnet,googlenet}, Object Detection \cite{RCNN}, etc., has  made features from well performing CNNs such as \cite{lenet}, Two-Stream CNN\cite{ma2017TS-LSTM}, 3DCNN \cite{3dcnn} ubiquitous for Action Recognition tasks including the zero shot setting. By using 3DCNN features in ZSL, a significant boost in accuracy has been observed \cite{wang2016zeroBidirectional}.
Semantic representation of a class provides additional, complementary information to the visual features of the classes. Typically, two types of semantic representations have been widely used in the ZSL literature: attribute representations \cite{attribute} and word vector representations \cite{word2vec}. Attribute representations are manually annotated vectors for each class based on the gesture and motion appearance of the objects in the video.  Word Vector representations are automatically learned from a large amount of textual data (Wikipedia Corpus). Word2vec models have been used successfully for extracting semantic word vectors from class names \cite{wang2016zeroBidirectional,kodirov2015unsupervisedDA,xu2016multitask-zero}. The core step in ZSL is to find a function or projection matrix which can establish a relationship between visual space and semantic space in such a way that visual features of classes map close to their semantic features and vice versa.  For example, we would like to have visual features of `running' map close to semantic features of `running' and far away from an unrelated action such as `eating'.

Note that our framework is similar in spirit to such methods with a key difference: Instead of learning a mapping between the semantic feature and visual features, we learn a mapping from the semantic features and the parameters of the \emph{distributions} representing the classes.

Most methods for zero-shot learning are evaluated on image classification whereas only a few methods have been proposed for zero-shot action recognition in the literature \cite{wang2016zeroBidirectional,xu2016multitask-zero,xu2017transductive,qin2017zero}. Such methods typically assume the inductive or the transductive setting. The most popular approach to ZSL is learning a linear compatibility between the visual and semantic space  \cite{SJE}. \cite{ESZSL2015,SAE2017} provide novel regularizations while learning a linear compatibility function.  ESZSL \cite{ESZSL2015} models the relationship between features and attributes as a linear compatibility function while explicitly regularizing the objective. UDA \cite{UDA} uses a domain adaptation technique by using unlabeled data of unseen classes for better estimation of the parameters.

Our model is inspired by the recently proposed model~\cite{verma2017simple}, which is a simple generative approach for zero-shot learning. However, their model does not have the reconstruction regularizer (autoencoder-style reverse mapping) from visual to attribute space and their focus is on image classification whereas here we have focused on action recognition. In another recent work, \cite{SAE2017} proposed a semantic auto-encoder for zero-shot learning which introduced the reconstructability regularizer. This paper works only in the inductive setting and their approach is not generative. Our generative approach can be seen as a combination of the generative approach of \cite{verma2017simple} with auto-encoder style regularizer proposed by \cite{SAE2017}. 

Among prior works on zero-shot action recognition in transductive setting, \cite{xu2016multitask-zero} proposed a transductive framework for zero-shot action recognition, which uses unlabeled unseen class data for training the model. In their work, they introduced a manifold-regularized regression and a data augmentation strategy to enhance the performance. They have also introduced a multi-task visual-semantic mapping for zero-shot action recognition.  In addition, they used prioritized auxiliary data augmentation for domain adaptation and improved the mapping between visual and semantic spaces. 

Because of the generative nature of our proposed approach, we can \emph{synthesize} the data from unseen class based on attribute and train the classifier. This approach helps to reduce the baisness in the case of Generalize Zero-Shot Learning. The efficacy of the proposed approach for the GZSL as well as ZSL can be seen from the experiment on three standard datasets. 
\section{Experiments \label{sec:experiments}}

{\bf Datasets and Settings:} 
We evaluate our proposed method in three of the most challenging video action recognition datasets, UCF101 \cite{ucf101soomro2012}, HMDB51 \cite{hmdb51kuehne2011} and Olympic \cite{Olympic}, widely used as benchmark datasets. We report mean accuracy along with standard deviation on 30 independent test runs with random train/test class splits. 
\begin{itemize}
\item {\bf UCF101:} \cite{ucf101soomro2012} is human action recognition data set with 101 different classes of actions and total of 13320 video clips. In our experiments, we split the classes into 51 seen and 50 unseen class respectively.
`\item {\bf HMDB51:} \cite{hmdb51kuehne2011} is the one of the most challenging human action recognition dataset with 51 different classes of human actions and total number of 6766 video clips. Each class has more than 100 video clips. For the evaluation of our model, we perform a 26/25 split for seen and unseen classes respectively.
\item {\bf Olympic:} \cite{Olympic} This dataset has 783 videos from 16 different classes with seen/unseen class split being 8/8. 
\vspace{-2.5em}
\end{itemize}
\begin{table}[H]
\scriptsize
\addtolength{\tabcolsep}{-2.0pt}
\begin{centering}
\begin{tabular}{|c|c|c|c|c|}
\hline 
\textbf{Dataset} & \textbf{\#videos} &\textbf{\#classes} & \textbf{seen/unseen}& \textbf{Attribute dim} \tabularnewline
\hline 
\hline 
UCF101 & 13320 & 101& 51/50 & 115 \tabularnewline
\hline 
HMDB51 &6676  & 51& 26/25& N/A \tabularnewline
\hline 
Olympic & 783 & 16& 8/8& 40 \tabularnewline
\hline 
\end{tabular}
\vskip 7pt
\caption{Dataset details and their train test split on all the three dataset used in our experiment.}
\par\end{centering}
\end{table}
\vspace{-1em}
{\bf Visual features:}
The quality of visual features directly affect the efficacy of the model. We use deep features as they have been shown to be successful in many  computer vision tasks.  In our experiments, we use the latest  convolutional 3D(C3D) visual features provided by \cite{tran2015learning}. This model was pre-trained on the sports-1M dataset. We extract the outputs of fc6 layer for all segments similar to \cite{tran2015learning} and then averaged over the segments to form a 4096-dimensional video representation which is used as the input visual features.

\begin{table*}[t]
\small
\addtolength{\tabcolsep}{24pt}
\centering{}%
\begin{tabular}{|c|c|c|c|c|}
\hline 
\textbf{Method} & \textbf{Embed} & \textbf{Olympic} & \textbf{UCF101} & \textbf{HMDB51} \tabularnewline
\hline 
\hline 
HAA \cite{HAA} & A & 46.1 $\pm$ 12.4 & 14.9 $\pm$ .8 & N/A \tabularnewline
DAP \cite{DAP} & A & 45.4 $\pm$ 12.8 & 14.3 $\pm$ 1.3 & N/A \tabularnewline
IAP \cite{IAP} & A & 42.3$\pm$12.5 & 12.8 $\pm$ 2& N/A \tabularnewline
ST \cite{ST} & W & N/A & 13.0$\pm$2.7 & 10.9$\pm$1.5 \tabularnewline
SJE \cite{SJE} & W & 28.6$\pm$4.9 & 9.9$\pm$1.4  & 13.3$\pm$2.4 \tabularnewline
SJE \cite{SJE} & A & 47.0$\pm$14.8 & 12.0$\pm$1.2  & N/A \tabularnewline
ESZSL \cite{ESZSL2015} & W & 39.6$\pm$9.6 & 15.0$\pm$1.3 & 18.5$\pm$2 \tabularnewline
UDA \cite{UDA} & A & N/A & 13.2$\pm$1.9 & N/A \tabularnewline
Bi-dir \cite{wang2016zeroBidirectional} & A & N/A & 20.5$\pm$.5 & N/A \tabularnewline
Bi-dir \cite{wang2016zeroBidirectional} & W & N/A & 18.9$\pm$.4 & 18.6$\pm$.7 \tabularnewline
\hline 
\textbf{Ours}& A & \textbf{50.41$\pm$11.2} & \textbf{22.74$\pm$1.2} & N/A  \tabularnewline
\textbf{Ours}& W & 34.12$\pm$10.1 & 17.33+1.1 & \textbf{19.28$\pm$2.1}  \tabularnewline
\hline 
\end{tabular}
\vskip 7pt
\caption{Results on inductive setting for standard zero shot learning setting(disjoint setting) for the action recognition. Here A represents the human annotated attribute vectors and W represents the \textit{word2vec} embedding. }
\label{ind}
\vspace{-1em}
\end{table*}

{\bf Class attributes:}
Two types of class attribute vectors (semantic representation of the classes) are widely used in ZSL: human labeled attributes \cite{attribute} and automatically learned distributed semantic representations such as word vectors \cite{word2vec}. Word vector representation is learned automatically by a skip-gram model  trained on the google news text corpus provided by Google. Each word is represented by a 300 dimensional vector. We experiment on both attribute and word2vec representations. For HMDB51 dataset, to the best of our knowledge, there is no publicly available attribute representations of the classes. Hence only word2vec is used for HMDB51. However, for UCF101 and Olympic datasets, 115 and 40 dimensional attribute vectors are available respectively \cite{ucf101soomro2012,Olympic}.

\vspace{-3em}
\noindent{\bf Hyper-parameters:}
Our model consists of four hyper-parameters: $\lambda_{\mu}$, $\lambda_{1}$ (Eq.~\ref{objectivefun}) and $\lambda_{\sigma^{2}}$, $\lambda_{2}$  (Eq.~ \ref{sigma_opt}) for estimating the projection matrix for mean  and variance. The optimal values of hyper-parameters are chosen via cross validation on the seen classes. For cross validation, we randomly fix 1/4th of the seen classes as validation classes and conduct five trials on 30 random splits (same as \cite{wang2016zeroBidirectional}). For generalized ZSL setting,  the number of synthesize examples for unseen classes is also hyper-parameter which we find using cross-validation and observe best model performance for 200 synthesized examples. 

\subsection{Inductive and Transductive ZSL}
In our first set of experiments, we evaluate our model for zero-shot action recognition with inductive and transductive setting and compare with a number of state-of-the-art methods.\\\\
{\bf Evaluation Metric:}
We evaluate our model using 30 different splits into seen and unseen classes provided by \cite{wang2016zeroBidirectional} for UCF101~(51/50), HMDB51~(26/25) datasets. For Olympic dataset, we generate  30 random splits for seen and unseen classes~(8/8). We use the average accuracy for all 30 splits as the evaluation metric. For fair comparison, we run five such trials for 30 random splits and present the final accuracy with average and standard deviation.\\
For generalized zero-shot setting we have evaluated for 30 different splits as above and calculated the average accuracy for seen and unseen classes. The final evaluation metric of our model is on the harmonic mean of the average accuracy of seen and unseen classes, as used in \cite{ConSE,HAA,SJE}.\\\\     
{\bf Inductive setting:} In this setting, it is assumed that only the labeled data from the seen classes is available during training. Table \ref{ind} shows the experimental results in the inductive setting of the zero-shot action recognition problem. We assume that the train and test classes are disjoint. Note that this assumption is made for all the evaluation settings in this work. In this setting, we obtain an improvement of \textbf{3\%} over the state-of-the-art on the Olympic dataset. On UCF-101, which is the most used dataset for zero shot action recognition, the proposed model outperforms state-of-the-art on attribute-based semantic representations. For HMDB dataset, the attribute vectors are not available. Hence, we present results only on word2vec embeddings.  Our model outperforms the state-of-the-art for this dataset as well. We believe the improvements can be attributed to its inherent nature of sharing information across classes (by modeling each as a basis combination of prototype classes) and its simple estimation procedure. 

\begin{table}[!htbp]
\small 
\addtolength{\tabcolsep}{-2.0pt}
\centering{}%
\begin{tabular}{|c|c|c|c|c|}
\hline 
\textbf{Method} & \textbf{Embed} & \textbf{Olympic} & \textbf{UCF101} & \textbf{HMDB51} \tabularnewline
\hline 
\hline 
 PST \cite{PST}& A & 48.6$\pm$11 & 15.3 $\pm$2.2& N/A \tabularnewline
ST \cite{ST} & W & N/A & 15.8$\pm$2.3 & 15.0$\pm$3 \tabularnewline
 TZWE \cite{xu2017transductive} & A & 53.5$\pm$11.9 & 20.2$\pm$2.2 & N/A \tabularnewline
TZWE \cite{xu2017transductive} & W & 38.6$\pm$10.6 & 18.0$\pm$2.7 & 19.1 $\pm$3.8 \tabularnewline
Bi-dir \cite{wang2016zeroBidirectional} & A & N/A &\textbf{28.3$\pm$1.0} & N/A \tabularnewline
Bi-dir \cite{wang2016zeroBidirectional} & W & N/A & 21.4$\pm$.8 & 18.9$\pm$1.1 \tabularnewline
UDA \cite{UDA} & A & N/A & 13.2$\pm$.6 & N/A \tabularnewline
\hline 
\textbf{Ours}& A & \textbf{57.88$\pm$14.1} & 24.48$\pm$2.9 & N/A \tabularnewline
\textbf{Ours}& W & 41.27$\pm$11.4 & 20.25$\pm$1.9 & \textbf{20.67$\pm$3.1}\tabularnewline
\hline 
\end{tabular}
\vskip 7pt
\caption{Results on transductive setting for the standard zero shot action recognition. Here A represents the human annotated attribute vectors and W represents the \textit{word2vec} embedding.}
\label{transd}
\end{table}

\noindent {\bf Transductive setting:} In the transductive setting, it is assumed that the unlabeled data of the unseen classes is also available at train time. Table \ref{transd} shows the performance of our model in the transductive setting. The unlabeled data from unseen classes helps us mitigate the bias towards the seen classes. In this setting, our model outperforms the state-of-the-art in the Olympic and HMDB datasets. The performance on the UCF-101 dataset is slightly worse, where \cite{wang2016zeroBidirectional} has the best performance. However, note that we outperform \cite{wang2016zeroBidirectional} in the inductive setting. \\ \\

\vspace{-3em}
\subsection{Generalized ZSL}
In this setting, the test data may come from both seen and unseen classes. In this setting, from the seen classes, we separate 20$\%$ of the data for testing and remaining 80$\%$ data is used as training data for calculating $\mathbf{W}_{\mu}$ and $\mathbf{W}_{\sigma^2}$ which is used to predict the mean ($\bm{\mu}_c$) and variance ($\bm{\sigma}_c^2$) for the unseen classes. One way to handle this setting is to assign each test data-point to the class whose estimated distribution gives the highest score. However, we notice that such an approach is biased towards seen classes since the model has not seen any unseen class examples.  In our approach, we propose the following solution to this issue: we synthesize class instances of unseen classes using the $\bm \mu_c$ and $\bm \sigma_{c}^{2}$ which are obtained from the transductive setting approach; these class instances are called pseudo class instances for unseen classes. Here we generate 200 instances for each unseen classes. Since we now have labelled data for seen classes and pseudo labelled data for unseen classes, we train SVM classifier for labelled seen classes data and pseudo labelled data for unseen classes. We then pass the test data (unseen class data plus $20\%$ seen class data) to the trained SVM classifier for classification. 
Table \ref{gentrad} presents the performance of our model in the generalized setting for zero-shot action classification which clearly shows that it significantly outperforms state-of-art on all the datasets. 
\begin{table}
\small
\addtolength{\tabcolsep}{-2.0pt}
\centering{}%
\begin{tabular}{|c|c|c|c|c|}
\hline 
\textbf{Method} & \textbf{Embed} & \textbf{Olympic} & \textbf{UCF101} & \textbf{HMDB51} \tabularnewline
\hline 
\hline 
HAA \cite{HAA} & A & 49.4 $\pm$ 10.8 & 18.7 $\pm$ 2.4 & N/A \tabularnewline
 SJE \cite{SJE} & W & 32.5$\pm$6.7 & 8.9$\pm$2.2  & 10.5$\pm$2.4 \tabularnewline
 ConSE \cite{ConSE} & W & 37.6 $\pm$ 9.9 & 12.7 $\pm$ 2.2 & 15.4$\pm$ 2.8 \tabularnewline
\hline 
\textbf{Ours}& A & \textbf{52.41$\pm$12.2} & \textbf{23.74$\pm$1.2} & N/A  \tabularnewline
\textbf{Ours}& W & \textbf{42.23$\pm$10.2} &  \textbf{17.45$\pm$2.2} & \textbf{20.10$\pm$2.1}  \tabularnewline
\hline 
\end{tabular}
\vskip 7pt
\caption{Results on the transductive setting for generalized zero-shot learning setting for the  action recognition. Here A represents the human annotated attribute vectors and W represents the \textit{word2vec} embedding. }
\label{gentrad}
\end{table}

\subsection{Few-shot action recognition}
\begin{figure}[H]
\begin{centering}
\includegraphics[scale=0.25]{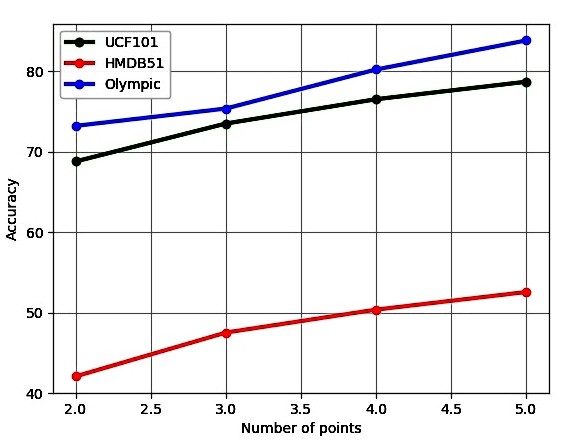}
\caption{Accuracy vs number of data points for few-shot learning }
\label{fewshot-figure}
\par\end{centering}
\vspace{-1em}
\end{figure}

Finally, we experiment with the few shot action recognition setting and present the results. Here only a small number of examples for each of the unseen classes are available during training. Our generative model provides a simple way to update the parameters of the class distribution using equation \ref{fs1}, \ref{fs2} . It is clear from the Table \ref{few-shotin} that availability of the few data points of the unseen classes significantly improves the performance which is now comparable to that of supervised learning.
Note that we do not assume any unlabeled data from the unseen classes in this setting. We test our model with varying number of examples of each unseen classes. The plot of accuracy with respect to the number of samples per class is shown in Figure \ref{fewshot-figure}. 


\begin{table}[H]
\scriptsize
\addtolength{\tabcolsep}{-2.0pt}
\centering{}%
\begin{tabular}{|c|c|c|c|c|}
\hline 
\textbf{Dataset} & \textbf{2 samples} & \textbf{3 samples} & \textbf{4 samples} & \textbf{5 samples}\tabularnewline
\hline 
\hline 
\textbf{UCF101} & 68.78$\pm$3.3 & 73.49$\pm$2.2 & 76.51$\pm$2.1 & 78.68$\pm$1.8 \tabularnewline
\hline 
\textbf{HMDB51} & 42.10$\pm$3.6 & 47.54$\pm$3.3 & 50.34$\pm$3.4 & 52.58$\pm$3.1 \tabularnewline
\hline 
\textbf{Olympic} & 73.20$\pm$7.4 & 75.35$\pm$7.3 & 80.21 $\pm$7.24& 83.81$\pm$7.11 \tabularnewline
\hline 
\end{tabular}
\vskip 7pt
\caption{Inductive setting with few-shot action recognition}
\label{few-shotin}
\end{table}
\vspace{-2em}

\section{Conclusion}\label{sec:conclusions}
We have presented a simple, probabilistic, generative model based framework for zero-shot action recognition. The proposed approach performs well in both the inductive and transductive setting for the standard (disjoint) and generalized zero-shot learning. The generative aspect of our model unables synthesizing unseen class examples and can effectively work in the generalized ZSL setting. In addition, the ability of leverage unlabeled data (transductive setting) helps address the domain shift problem between seen and unseen classes. A particularly appealing aspect of our model is that it yields a closed form solution for the parameters to make it fast and easy to implement. Experimental results are shown to achieve state-of-the-art performance. The proposed method also generalizes to few-shot action recognition setting, achieving comparable results to \textit{fully supervised} learning using only a few synthesized examples from each unseen class.

\small{\textbf{Acknowledgments:} Thanks to Vismay who always gives his support in discussion and experiment. Special thanks to Prof. Hema A Murthy for her valuable feedback. Vinay Verma acknowledges support from Visvesvaraya Ph.D. fellowship. Piyush Rai acknowledges support from Visvesvaraya Faculty Fellowship and a grant from Tower Research.}

{\small
\bibliographystyle{ieee.bst}
\bibliography{egbib}
}

\end{document}